\documentclass[10pt,twocolumn,letterpaper]{article}

\usepackage{wacv}
\usepackage{times}
\usepackage{epsfig}
\usepackage{graphicx}
\usepackage{amsmath}
\usepackage{amssymb}
\usepackage{bm}
\usepackage{algorithm}
\usepackage{algpseudocode}
\usepackage{subcaption}
\usepackage{varwidth}
\usepackage{scalerel}
\usepackage[accsupp]{axessibility}  

\wacvfinalcopy 

\ifwacvfinal
\def\assignedStartPage{1} 
\fi

\ifwacvfinal
\usepackage[breaklinks=true,bookmarks=false]{hyperref}
\else
\usepackage[pagebackref=true,breaklinks=true,colorlinks,bookmarks=false]{hyperref}
\fi

\usepackage{accents}
\newlength{\dhatheight}
\newcommand{\doublehat}[1]{%
    \settoheight{\dhatheight}{\ensuremath{\hat{#1}}}%
    \addtolength{\dhatheight}{-0.35ex}%
    \hat{\vphantom{\rule{1pt}{\dhatheight}}%
    \smash{\hat{#1}}}}

\ifwacvfinal
\else
\fi

\begin{document}

\title{Learnable Adaptive Cosine Estimator (LACE) for Image Classification}

\author{Joshua Peeples \quad Connor H. McCurley \quad Sarah Walker \quad Dylan Stewart \quad Alina Zare\\
Department of Electrical \& Computer Engineering, University of Florida\\
{\tt\small \{jpeeples, cmccurley, sarah.walker, d.stewart, azare\} @ufl.edu}
}

\maketitle

\begin{abstract}
In this work, we propose a new loss to improve feature discriminability and classification performance. Motivated by the adaptive cosine/coherence estimator \cite{Scharf1996ACE} (ACE), our proposed method incorporates angular information that is inherently learned by artificial neural networks.  Our learnable ACE (LACE) transforms the data into a new ``whitened" space that improves the inter-class separability and intra-class compactness.  We compare our LACE to alternative state-of-the art  softmax-based and feature regularization approaches.  Our results show that the proposed method can serve as a viable alternative to cross entropy and angular softmax approaches. Our code is publicly available.\footnote{\url{https://github.com/GatorSense/LACE}} 
\end{abstract}
\section{Introduction}
\label{sec:introduction}

Feature embedding models for image classification should map data from the same class onto a single point that is metrically far from embeddings of alternative classes \cite{Bengio2014RepLearningReview, Murphy2012ProbabilisticML, Bishop2006PatternRec, Theodoridis2008PatternRec, McCurley2021ResearchProposal}.  State-of-the-art (SOA) classification networks typically use softmax or embedding losses to guide this feature learning \cite{Schroff2015FaceNet, He2015ResNet,Krizhevsky2012AlexNet,Simonyan2015VGGNet,Szegedy2015InceptionNet}.  Yet, \textit{softmax loss} \cite{Taigman2014DeepFace,Sun2014DeepLearningFace} has no inherent property to encourage intra-class compactness or inter-class variation and \textit{embedding loss} \cite{Schroff2015FaceNet,Hermans2017DefenseTripletLoss, Koch2015SiameseNetworks, Koch2015SiameseNetworksThesis} methods have proven difficult and untimely to train due to the necessity of hard mining schemes \cite{Wen2016DiscFeatureLearning,Liu2017LargeMarginSoftmax,Liu2017SphereFace,Hermans2017DefenseTripletLoss,Schroff2015FaceNet}.  Consequentially, many studies have investigated variants of both softmax and embedding losses to encourage better discriminating performance \cite{Wang2018CosFace}. 

The works of \cite{Wen2016DiscFeatureLearning,Chopra2005LearninSimilarityMetric,Hadsell2006DimReduction,Hoffer2015DeepMetric,Wang2014DeepRanking,Liu2017LargeMarginSoftmax,Qi2017CenterLoss} adopt multi-loss learning to increase feature discriminability.  While these approaches increase classification performance over traditional softmax loss, they fail to promote both intra-class compactness and inter-class separability.  Several works attempt to utilize joint supervision by combining softmax with an auxiliary, Euclidean-based margin loss \cite{Sun2014JointIdentification,Sun2015Sparsifying,Wen2016DiscFeatureLearning}.  Feature representations guided by softmax loss often show angular separability \cite{Liu2017SphereFace,Li2018AngularSoftmax,Wang2018CosFace,Deng2019ArcFace}. To maintain the intrinsic consistency of the features learned by softmax loss neural networks, it is instead favorable to use angular metrics \cite{Choi2020AMCLoss} to promote class compactness and separability as opposed to Euclidean-based approaches \cite{Schroff2015FaceNet,Koch2015SiameseNetworks,Chen2020ContrastiveLearning,Sohn2016NPairLoss}.  Angular softmax losses \cite{Liu2017SphereFace, Li2018AngularSoftmax,Wang2018CosFace,Deng2019ArcFace} were recently proposed to take advantage of this knowledge.  As with the previous approaches, these losses fail to promote both inter-class separability and intra-class compactness or rely heavily on hyperparameters which can be difficult to select \textit{a priori}.

\begin{center}
    \begin{figure*}[htb]
        \centering
        \includegraphics[width=1\textwidth]{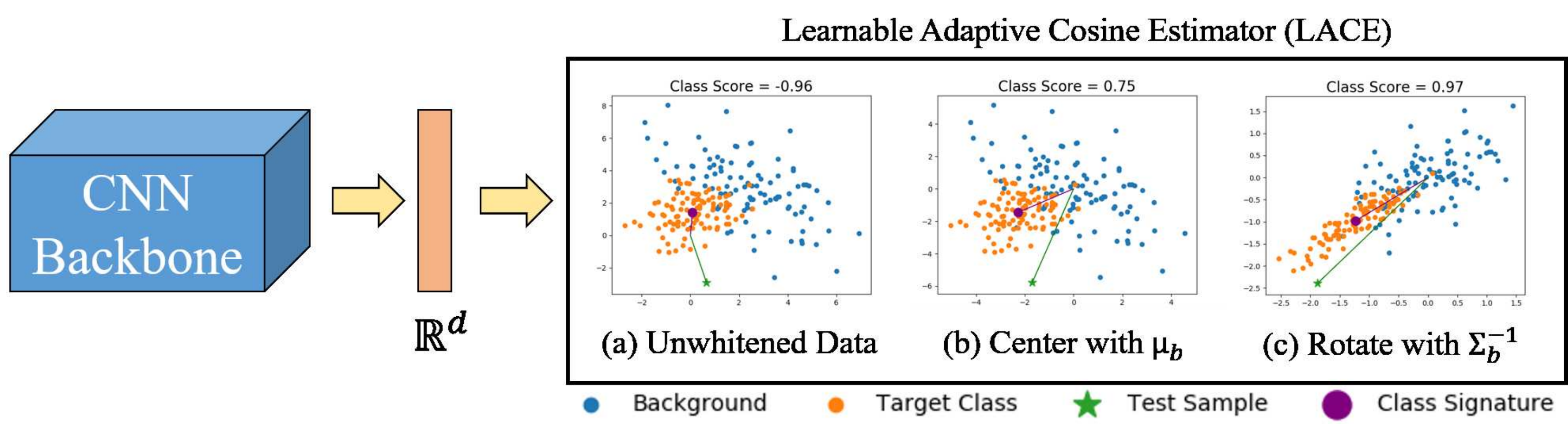}
        \caption[Learnable ACE Loss.]{Network with Learnable ACE (LACE) loss. The backbone architecture (\textit{e.g.}, ResNet18 \cite{He2015ResNet}) extracts convolutional feature maps and the features are aggregated (\textit{i.e.}, global average pooling) to be a $\mathbb{R} ^{d}$ feature vector. Here we show a toy example of the transformation performed in LACE. By subtracting the background mean ($\mu_b$) and multiplying by the inverse background covariance matrix ($\Sigma_b$), we can improve the estimated score for a new test sample. In LACE, the features are projected into a whitened data space where the $C$ target classes (orange) are compared with the non-target classes (blue). The LACE scores for each sample are then passed into the softmax function and the loss is computed as shown in Equation \ref{eq:LACE_objective}. Network parameters (\textit{i.e.}, target signatures, background mean, and background covariance) are updated via backpropagation. In Figure \ref{fig:neural_ace_loss_network}a, many background samples (\textit{i.e.}, samples not in target class) would obtain a high cosine similarity with the target class signature (\textit{i.e.}, class average in this example), while at least half of the true target samples would not. In Figure \ref{fig:neural_ace_loss_network}b, which shows background mean subtraction, the cosine similarity between the target signature and true targets drastically increases ($-0.96$ to $0.75$) while the similarity with background samples decreases. The cosine similarity between target samples and the target class signature is further improved post whitening rotation ($0.75$ to $0.97$) as shown in Figure \ref{fig:neural_ace_loss_network}c. Once whitening is completed, the background data should be centered and spherical (\textit{i.e.}, identity covariance and zero mean vector). In our figure, the background data may not appear spherical due to scale of axes.} 
        \label{fig:neural_ace_loss_network}
    \end{figure*}
\end{center}

In this paper, we introduce an alternative loss which adopts the discriminative adaptive cosine/coherence estimator (ACE) \cite{Zare2018MIACE,Ermolov2021Whitening,Su2021Whitening} to encourage angular separation among classes.  As illustrated in Figure \ref{fig:neural_ace_loss_network},  the \textit{Learnable ACE (LACE) loss} takes features from a model and whitens them according to a global background distribution.  The ACE detection statistic is computed between the samples and each of the $C$ class representatives in the whitened space.  Multi-class classification is performed by taking the softmax over ACE outputs. The proposed method demonstrates three key features: 1) it promotes both inter-class separability and intra-class compactness without the need of a separation hyperparameter, 2) unlike comparable methods, LACE enforces strong regularization constraints on the feature embeddings through a whitening transformation and 3) target representatives and data whitening statistics can be updated easily using backpropagation.  The proposed LACE loss complements the angular features inherently learned with softmax loss and can be used in joint supervision tasks. 


Our contributions can be summarized as follows:
\begin{itemize}
    \item Apply our whitening transformation based on ``background" information to improve feature discriminability - unlike previous approaches 
    \item Ability to learn target signatures and non-target statistics (\textit{i.e.}, background mean and covariance matrix) through backpropagation
    \item Outperform other angular regularization approaches without hyperparameters (\textit{i.e.}, scale and margin)
\end{itemize}

\section{Background and Related Work}
\label{sec:related_work}


\paragraph{Softmax}  Given a set of $N$ training input features $\mathbf{X} \in \mathbb{R}^{N \times D}$ for dimensionality $D$, a deep CNN will condense the data into a lower dimension $d$, (typically $512$ or $1024$), through an embedding function, $f$.  This gives embedding vectors $\mathbf{Z} = f(\bm{X}) \in \mathbb{R}^{N \times d}$, which are often passed to a fully-connected layer(s) for classification with cross-entropy + softmax loss or compared to alternative sample embeddings directly with an embedding loss \cite{Khan2020Survey, Kaya2019MetricLearningSurvey}. If $\hat{\mathbf{Y}} \in \mathbb{R}^{N \times C}$ are the sample predictions for $C$ classes, and $\mathbf{Y} \in \mathbb{R}^{N \times C}$ are the true class labels, then the \textit{cross-entropy loss} for class $c$ is given by 
\begin{equation} \label{eq:cross_entropy_loss_regular}
   L_{CE} = \frac{1}{N}\sum_{n=1}^{N}-\log \left ( \frac{\exp{{\hat{y}_{nc}}}}{\sum_{j=1}^{C}\exp{\hat{y}_{nj}}} \right ),  
\end{equation}
\noindent
where $\hat{y}_{nc}$ denotes the $c$-th activation of the vector of predicted class scores $\hat{\bm{y}}_{n}$ for sample embedding $\bm{z}_{n}$.  In the softmax loss, $\hat{\bm{y}}_{n}$ contains the activations of a fully connected layer $\bm{W} \in \mathbb{R}^{d \times C}$.  Thus, $\hat{\bm{y}}_{nc} = \bm{W}^{T}_{c}\bm{z}_{n} + b_{c}$, where $\bm{W}_{c} \in \mathbb{R}^{d}$ is the $c$-th column of weight matrix $\bm{W}$ and $b_{c}$ is a bias term. Because this activation is the result of a dot product, \cite{Liu2017LargeMarginSoftmax} showed that the softmax can be reformulated in terms of the angle $\theta_{c} (0 \leq \theta_{c} \leq \pi)$ between vectors $\bm{W}_{c}$ and $\bm{z}_{n}$.  As shown in Equation \ref{eq:cross_entropy_loss_angular}, the embedding features inferred by a deep CNN with softmax loss are encouraged to show angular distributions (similar to \cite{Liu2017LargeMarginSoftmax}, the bias term is omitted to simplify analysis): 

\begin{equation} \label{eq:cross_entropy_loss_angular}
\small
   L_{CE} = \frac{1}{N}\sum_{n=1}^{N} -\log \left ( \frac{\exp{\lVert\bm{W}_{c}\rVert \lVert\bm{z}_{n}\rVert cos(\theta_{c})}}{\sum_{j=1}^{C}\exp{\lVert\bm{W}_{j}\rVert \lVert \bm{z}_{n} \rVert cos(\theta_{j})}} \right ).
\end{equation}
Methods using cross entropy + softmax loss are widely used and obtain top scores for many image classification tasks, such as common object and land cover classification, object localization and face recognition \cite{Jia2021DeepLearningHSI, Deng2019ArcFace,Lu2007ImageClassificationSurvey,Sornam2017ImageClassificationSurvey,Nath2014ImageClassificationSurvey,Schmarje2021ImageClassificationSurvey,Touvron2020FixingA,Touvron2020FixingB}.

\paragraph{Angular Losses}
Motivated by the angular feature distributions shown by softmax loss \cite{Choi2020AMCLoss,peeples2020divergence}, angular margin losses have recently been utilized with much success \cite{Liu2017LargeMarginSoftmax,Deng2019ArcFace,Wang2018CosFace,Deng2020SubcenterArcFace}.  Large-margin softmax (L-Softmax) \cite{Liu2017LargeMarginSoftmax} was among the first to characterize softmax loss in terms of angles. L-Softmax promotes inter-class separability by enforcing an angular margin of separation on the classes.  Angular softmax (A-softmax) \cite{Liu2017SphereFace} improves upon L-Softmax by projecting the Euclidean features into an angular feature space and introducing a parameter for inter-class separation.  Large-margin cosine loss (LMCL) \cite{Wang2018CosFace} shares the same fundamental ideas of A-Softmax, but ensures that each class is separated by a uniform margin.  Additive angular margin loss \cite{Deng2019ArcFace} directly considers both angle and arc to promote inter-class separability on the normalized hypersphere.  Finally, angular margin contrastive loss \cite{Choi2020AMCLoss} utilizes training concepts commonly used with ranking losses to jointly optimizes an angular loss along with softmax loss.   Similarly to Euclidean-based embedding loss methods, angular losses have been utilized primarily in facial recognition tasks \cite{Deng2020SubcenterArcFace,Zhang2019AdaCos}. 

\paragraph{Data Whitening}
Whitening is a powerful transformation that is used to normalize data. Whitening consist of scaling, rotating, and shifting the data to achieve zero means, unit variances, and decorrelated features \cite{ioffe2015batch}.  An example of data whitening in deep learning models is batch normalization \cite{ioffe2015batch}. Batch normalization has been shown to not only improve model performance, but also training stability. Extensions to the baseline batch normalization approach have been proposed to further improve computational costs and performance of data whitening in deep neural networks \cite{huang2019iterative,luo2017learning,desjardins2015natural,zhang2021stochastic,huang2018decorrelated}. In addition to applying data whitening within neural networks, loss functions that incorporate the whitening transformation have been introduced to achieve better performance \cite{Ermolov2021Whitening}. However, all previous approaches apply the data transformation based on representatives from all the data (\textit{e.g.}, via mini-batch). To improve inter-class separation and intra-class compactness of the data samples, the whitening transform can be computed based on ``background" data \cite{Zare2018MIACE,Meerdink2020MTMIHSI}, or samples not in the target class, to improve data discriminability.  Our work investigates use of data whitening with background statistics on classification performance in deep neural networks.
\paragraph{ACE} Many approaches in binary classification are statistical methods in which the target, or class of interest, and background information (all other classes) are modeled as random variables distributed according to an underlying probability distribution \cite{Zare2018MIACE,Murphy2012ProbabilisticML,Theodoridis2008PatternRec}.  Thus, the classification problem can be posed as a binary hypothesis test: target absent ($\bm{H}_{0}$) or target present ($\bm{H}_{1}$), and a classifier can be designed using the generalized likelihood ratio test (GLRT) \cite{Vincent2020NonZeroACE,Manolakis2013ACE}.  The adaptive cosine/coherence estimator (ACE) \cite{Scharf1996ACE} is a popular and effective method for performing statistical binary classification which can be formulated as the square-root of the GLRT, 
\begin{small}
\begin{equation} \label{eq:ace_statistic}
    D_{ACE}(\bm{x},\bm{s}) = \frac{\bm{s}^{T}\Sigma_{b}^{-1}(\bm{x}-\bm{\mu}_{b})}{\sqrt{\bm{s}^{T}\Sigma_{b}^{-1}\bm{s}} \sqrt{(\bm{x}-\bm{\mu}_{b})^{T}\Sigma_{b}^{-1}(\bm{x}-\bm{\mu}_{b})}  },
\end{equation}
\end{small}
\noindent where $\bm{x}$ is a sample feature vector and  $\bm{s}$ is an \textit{a priori} target class representative. The background class distribution is parameterized by the mean vector ($\bm{\mu}_{b}$) and inverse covariance ($\Sigma_{b}^{-1}$). The response of the ACE classifier, $D$, is essentially the dot product between a sample and known class representative in a whitened coordinate space.  Similarly to the mentioned angular margin approaches, ACE measures the angular or cosine similarity of a sample with a known class exemplar.  This can be considered as comparing the spectral shape of the feature vectors.  ACE, however, measures this discrepancy in a coordinate space where data is whitened by the statistics of the background class distribution.  While inherently binary, ACE can be utilized in a multi-class setting by forming $C$ one-versus-all classifiers.  By applying softmax over all classifier responses and taking the maximum output, our proposed method provides a hard class label for new test samples.

\section{Proposed Method}\label{sec:method}
A brief overview of the image classification method using LACE is shown in Figure \ref{fig:neural_ace_loss_network}. A CNN feature extraction backbone (\textit{e.g.}, ResNet18 \cite{He2015ResNet}) is used to extract features from the input images. The features are then aggregated (\textit{i.e.}, global average pooling) to provide a deep feature representation in the form of $d$ length vectors. The backbone architecture was jointly trained from scratch using LACE to encourage the deep features to adhere to our metric and not be biased towards the initial pre-training of ImageNet. The LACE score for each sample is then computed and passed into a softmax activation function to produce a predicted probability for each class.
In this way, training our approach encourages that feature embedding positions for an image are angularly similar to their corresponding class representative. 
\subsection{LACE Loss}\label{sec:lace_loss}
Inspired by ACE \cite{Scharf1996ACE,Zare2018MIACE} and cross entropy, our LACE objective function is the following, 
\begin{equation} \label{eq:LACE_objective}
      L_{LACE} = \frac{1}{B}\sum_{n=1}^{B}-\log \left(\frac{\exp{\doublehat{\bm{s}}_{c}^T\doublehat{\bm{x}_n}}}{{\sum_{j=1}^{C}}\exp{\doublehat{\bm{s}}}_{j}^T\doublehat{\bm{x}_n}} \right),  
    \end{equation}
\noindent where $B$ is the mini-batch size and $C$ is the number of classes.
\noindent
The inner product shown here is simply the ACE similarity which stems from an expansion of Equation \ref{eq:ace_statistic}.  This is derived as  $\doublehat{\bm{s}_c}= \frac{\hat{\bm{s}}}{\lVert \hat{\bm{s}} \rVert}$, $\doublehat{\bm{x}}= \frac{\hat{\bm{x}}}{\lVert \hat{\bm{x}} \rVert}$, where $\hat{\bm{s}}=\bm{D}^{-\frac{1}{2}}\bm{U}^{T}\bm{s}$ and $\hat{\bm{x}}=\bm{D}^{-\frac{1}{2}}\bm{U}^{T}(\bm{x}-\bm{\mu}_{b})$.  Here $\bm{U}$ and $\bm{D}$ are the eigenvectors and eigenvalues of the inverse background covariance matrix, $\bm{\Sigma}^{-1}_{b}$, respectively.  Thus, the ACE value is the cosine similarity between a test point, $\bm{x}$, and a class representative, $\bm{s}_c$, in a whitened coordinate space.  The objective shown in Equation \ref{eq:LACE_objective} maximizes the ACE similarity between a sample and its corresponding class representative. By computing the loss on the ACE similarity, we encourage the network to learn discriminative features that promote samples belonging to the same class to be ``near" one another in the whitened coordinate space.  Since ACE is generally used for binary classification, we extend LACE to include $C$ target signatures.  Similar to the weight matrix in a standard fully connected layer, LACE learns a signature matrix, $\bm{S} \in \mathbb{R}^{d \times C}$, that is used for multi-class classification.  

As can be observed, our LACE loss requires three components which can be learned during training: 1) background mean, 2) background covariance and 3) target class representatives. Unlike other angular softmax alternatives, we do not need to set hyperparameters such as scale and margin beforehand. Therefore, we randomly initialize the target class signature matrix, background mean, and background covariance. The parameters for our proposed LACE method are updated via backpropagation. The derivatives for 1) background mean 2) background covariance and 3) target class representatives are provided in the supplementary material. Pseudo-code for our approach is shown in Algorithm \ref{alg:LACE}. 
\begin{algorithm}[htb]
	\caption{Mini-batch training of LACE loss}
	\label{alg:LACE}
	\begin{algorithmic}[1]
	\Require {$\bm{X} \in \mathbb{R}^{N \times D}$ training inputs, $\bm{Y} \in \mathbb{R}^{N \times C}$ corresponding class labels, $\bm{Z} \in \mathbb{R}^{N \times d}$ deep feature embeddings of $\bm{X}$, $f_{\bm{\theta}}(\bm{x})$ CNN feature embedding model with parameters $\bm{\theta}$}, epochs $T$, mini-batch size $B$
	\For{$t=1$ to $T$} 
	\For{each minibatch $B$}
	\State \begin{varwidth}[t]{\linewidth}
	    \hskip\algorithmicindent{$\bm{Z}_{B} \gets f_
	    {\bm{\theta}}(\bm{X}_{B})$} \par
		\hskip\algorithmicindent {Whiten data, $\bm{Z}_{B}$, using $\bm{\mu}_{b}, \bm{\Sigma}_{b}$} \par
		\hskip\algorithmicindent {Compute $L_{LACE}$ according to Eq. \ref{eq:LACE_objective}} \par
		\hskip\algorithmicindent {Update $\bm{\theta}$, $\bm{\mu}_{b}, \bm{\Sigma}_{b}$, $\bm{S}$ using backpropagation}
	\end{varwidth}
	\EndFor 
	\EndFor 
	\State {\Return $f_{\theta}$}
	\end{algorithmic}
\end{algorithm}
\subsection{Implementation and Training Details}
Our code was implemented in Python $3.8$ with PyTorch $1.9$.  All models were trained for a max of $300$ epochs using the Adam optimizer with mini-batch size of $n=256$ ($n=128$ for backbone architecture experiments in Section \ref{sec:Ablation}) and maximum learning rate of $.001$. Early stopping was used to terminate training if the validation loss did not improve within $10$ epochs. We use the Adam momentum parameters of $\beta_{1}= .9$ and $\beta_{2}= .999$. Experiments were ran on two 2080 TI GPUs. For LACE, we enforced the inverse covariance matrix to be a positive semi-definite (PSD) matrix. To satisfy this constraint, we performed matrix multiplication between the background covariance matrix and its transpose. Singular value decomposition (SVD) was then performed on the resulting PSD outer product to compute the eigenvectors and eigenvalues ($\bm{U}$ and $\bm{D}$) of the background covariance matrix.

A notable advantage of our approach is that LACE can be implemented easily and efficiently if you consider a single, global background class.  In this scenario, the target class signature matrix $\bm{S}$ and whitening rotation matrix act as weight matrices while the background mean vector acts as a bias. Thus, LACE scores can be easily computed over all classes in the feed-forward portion of the neural network.

\section{Experimental Results}
\label{sec:experiments}

We investigated four image datasets: FashionMNIST \cite{xiao2017fashion}, CIFAR10 \cite{krizhevsky2009learning}, SVHN \cite{netzer2011reading}, and CIFAR100 \cite{krizhevsky2009learning}. For CIFAR100, we used the coarse (\textit{i.e.}, $20$ classes) and sparse (\textit{i.e.}, $100$ classes) labels. We evaluated each comparison approach quantitatively through test classification accuracy. We held 10\% of the training data for validation and applied each trained model to the holdout test set. We performed three runs of the same random initialization for each method. The CNN backbone was trained from scratch for each loss function to promote the features learned by the model to adhere to the chosen metric. The average test classification accuracy is reported across the different runs. We followed the same training procedure and data augmentation as Xue et. al. \cite{xue2018deep}. The proposed method was compared to several SOA angular softmax and feature regularization approaches: arcface \cite{Deng2019ArcFaceAngularMarginLoss,Deng2019ArcFace}, cosface \cite{Wang2018CosFace}, sphereface \cite{Liu2017SphereFace}, center loss \cite{Wen2016DiscFeatureLearning}, and angular margin contrastive (AMC) loss \cite{Choi2020AMCLoss}. The hyperparameters for the angular margin ($m$) and scale ($\gamma$) for comparison approaches were set to the following values: arcface ($m=0.05$, $\gamma=30$), cosface ($m=0.4$, $\gamma=30.0$), and sphereface ($m=64.0$, $\gamma=1.35$). For AMC and center loss, the weighting terms on the constrastive loss was set to $.1$ and $.9$ on the softmax loss (the weights on both terms were constrained to sum to $1$). The backbone architecture for the SOA comparisons was ResNet18 \cite{He2015ResNet}.

\subsection{Comparison with SOA Approaches}
\label{sec:SOA}
\begin{table*}[htb!]
\centering
\caption{Test classification accuracy on each dataset for softmax, angular softmax alternatives, feature regularization methods, and our proposed approach (LACE) with ResNet18 as the backbone architecture. The best average result is bolded and the second best result is underlined.} 
\begin{tabular}{|c|c|c|c|c|c|}
\hline
Method     & CIFAR10        & FashionMNIST   & SVHN   & CIFAR100 (Coarse) & CIFAR100 (Sparse)       \\ \hline
Softmax    & 88.69$\pm$0.21 & 93.96$\pm$0.08 & 94.68$\pm$0.15 &76.38$\pm$0.76 &  66.86$\pm$0.76\\ \hline
Arcface \cite{Deng2019ArcFace}    & 88.75$\pm$0.27 & 93.71$\pm$0.55 & 94.93$\pm$0.32 &75.78$\pm$1.10 & 66.94$\pm$0.80 \\ \hline
Cosface \cite{Wang2018CosFace}   & 88.84$\pm$0.23 & \underline{94.28$\pm$0.04} & \underline{95.28$\pm$0.13} &77.04$\pm$0.40 & \underline{68.84$\pm$0.08}\\ \hline 
Sphereface \cite{Liu2017SphereFace} & 88.80$\pm$0.51 & 94.17$\pm$0.16 & 94.64$\pm$0.47 &75.74$\pm$0.37 & 65.12$\pm$0.78\\ \hline
AMC \cite{Choi2020AMCLoss} & 88.40$\pm$0.51 & 94.09$\pm$0.30 & 94.98$\pm$0.23 & \underline{77.20$\pm$0.28} & 67.68$\pm$0.22\\ \hline
Center Loss \cite{Wen2016DiscFeatureLearning} & \underline{89.02$\pm$0.39} & \textbf{94.29$\pm$0.09} & 94.75$\pm$0.50 & \textbf{77.77$\pm$0.10} & \textbf{70.03$\pm$0.90}\\ \hline
LACE (ours)     & \textbf{90.26$\pm$0.19} & 94.27$\pm$0.13 & \textbf{95.69$\pm$0.20} &63.32$\pm$1.00 & 33.12$\pm$0.56    \\ \hline
\end{tabular}
\label{tab:CE_outputs}
\end{table*}

\begin{figure*}[htb]
\centering
	\begin{subfigure}{.23\textwidth}{
			\includegraphics[width=\textwidth]{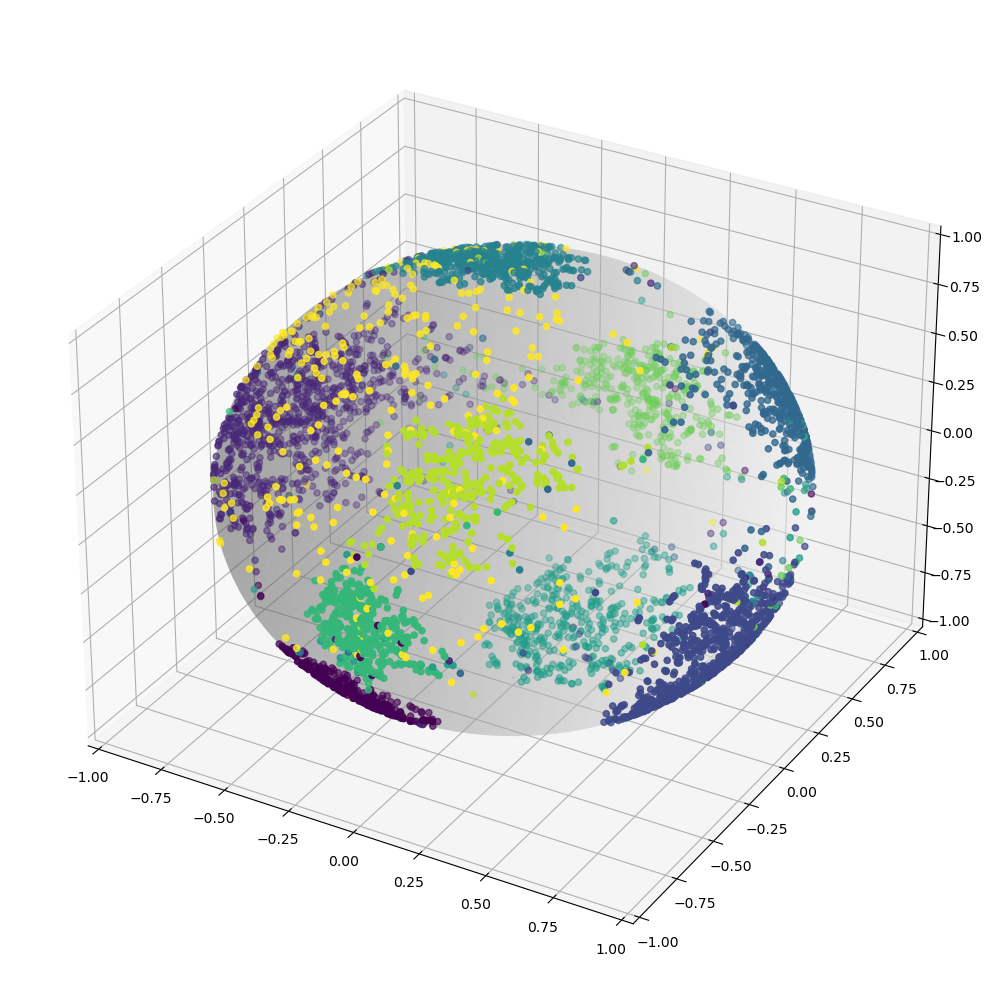}
			\caption{Softmax}
			\label{fig:Softmax}
		}
	\end{subfigure} 
	\begin{subfigure}{.23\textwidth}{
			\includegraphics[width=\textwidth]{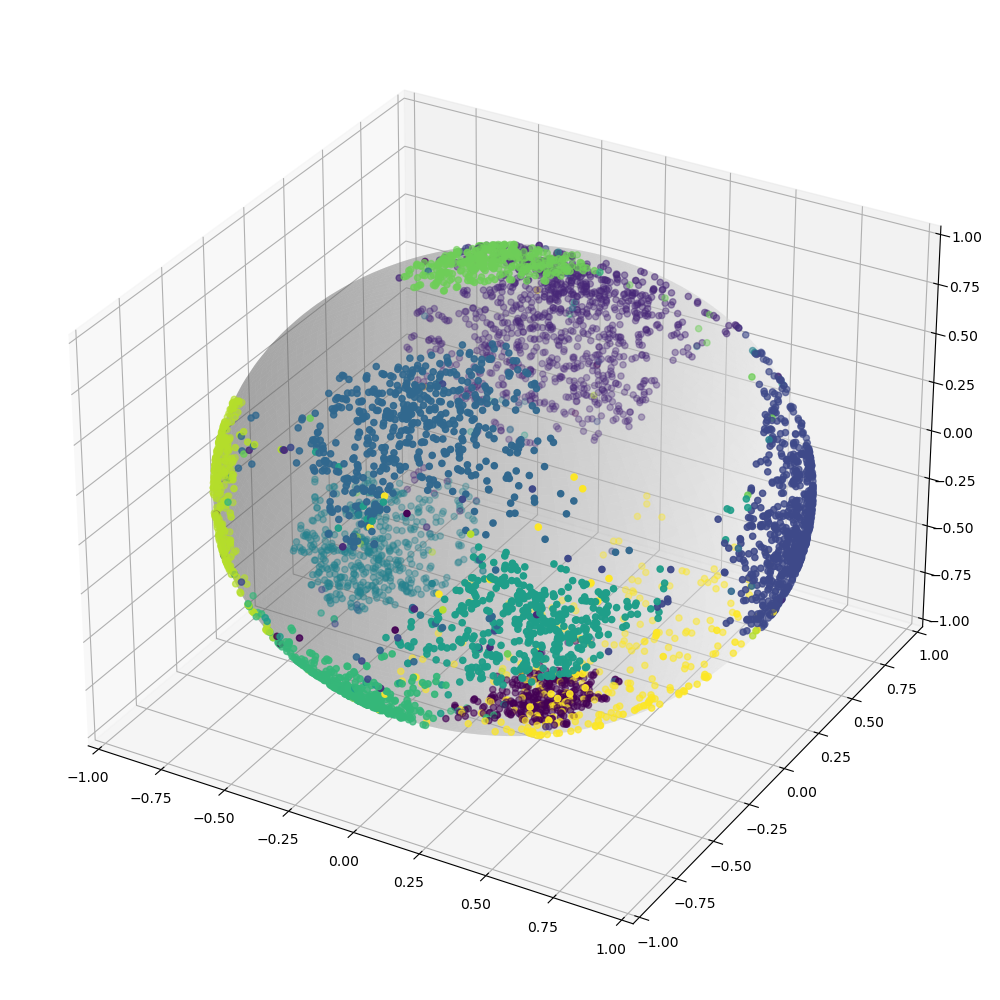}
			\caption{Arcface}
			\label{fig:Arcface}
		}
	\end{subfigure}
	\begin{subfigure}{.23\textwidth}{
			\includegraphics[width=\textwidth]{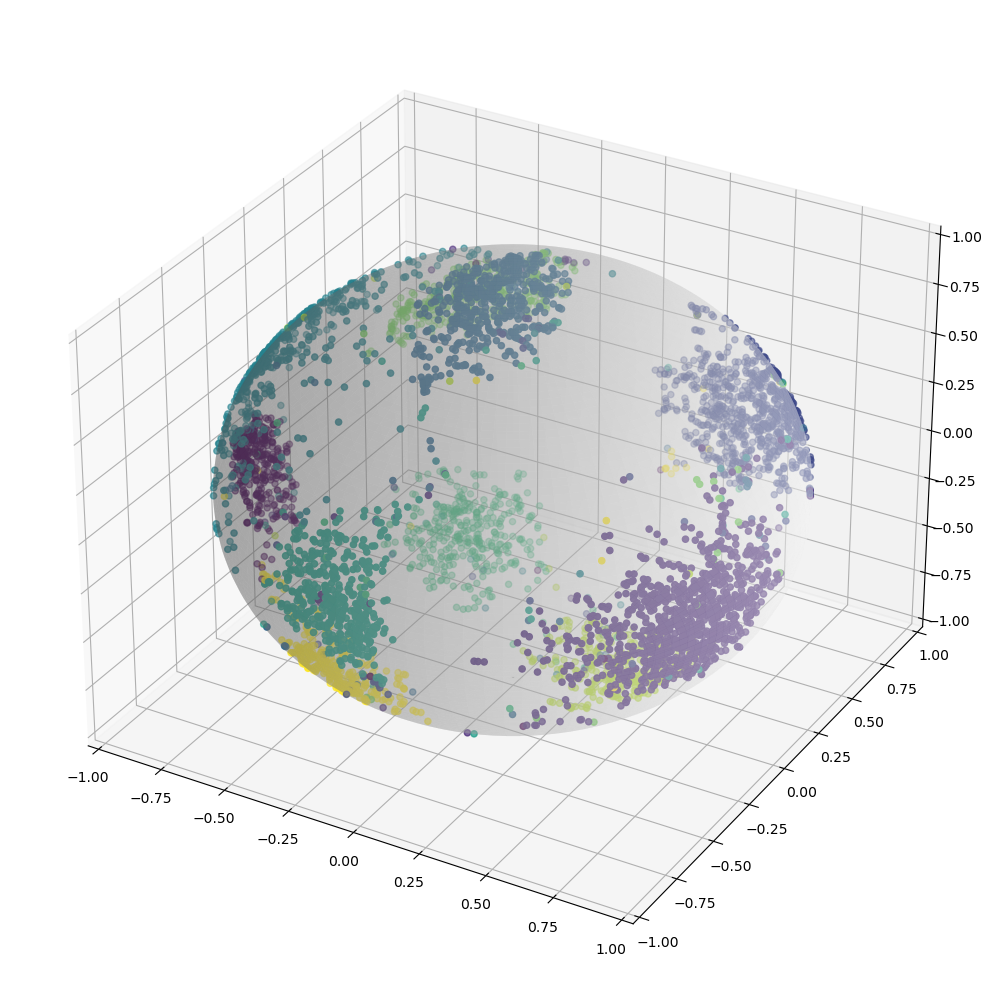}
			\caption{Cosface}
			\label{fig:Cosface}
		}
	\end{subfigure} 
	\begin{subfigure}{.23\textwidth}{
			\includegraphics[width=\textwidth]{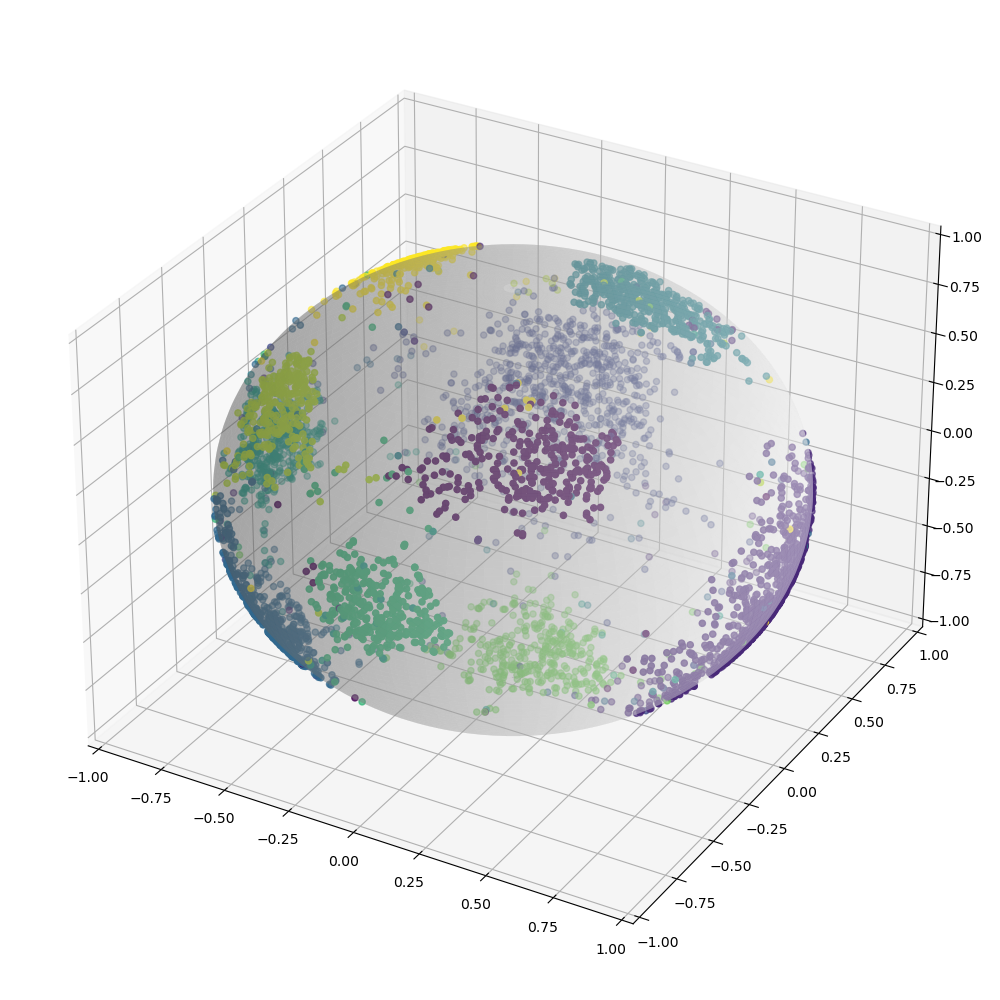}
			\caption{Sphereface}
			\label{fig:Sphereface}
		}
	\end{subfigure}
	
	\begin{subfigure}{.23\textwidth}{
			\includegraphics[width=\textwidth]{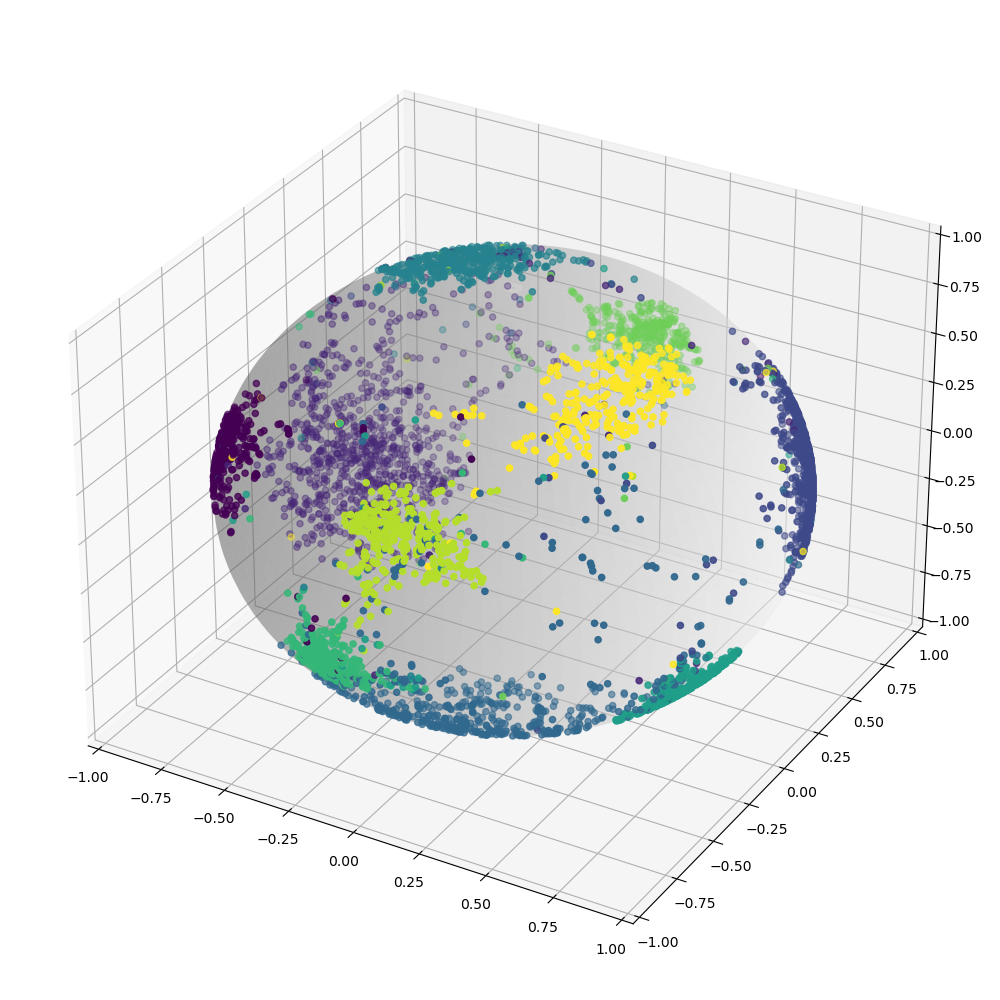}
			\caption{Center Loss}
			\label{fig:Center}
		}
	\end{subfigure}
	\begin{subfigure}{.23\textwidth}{
			\includegraphics[width=\textwidth]{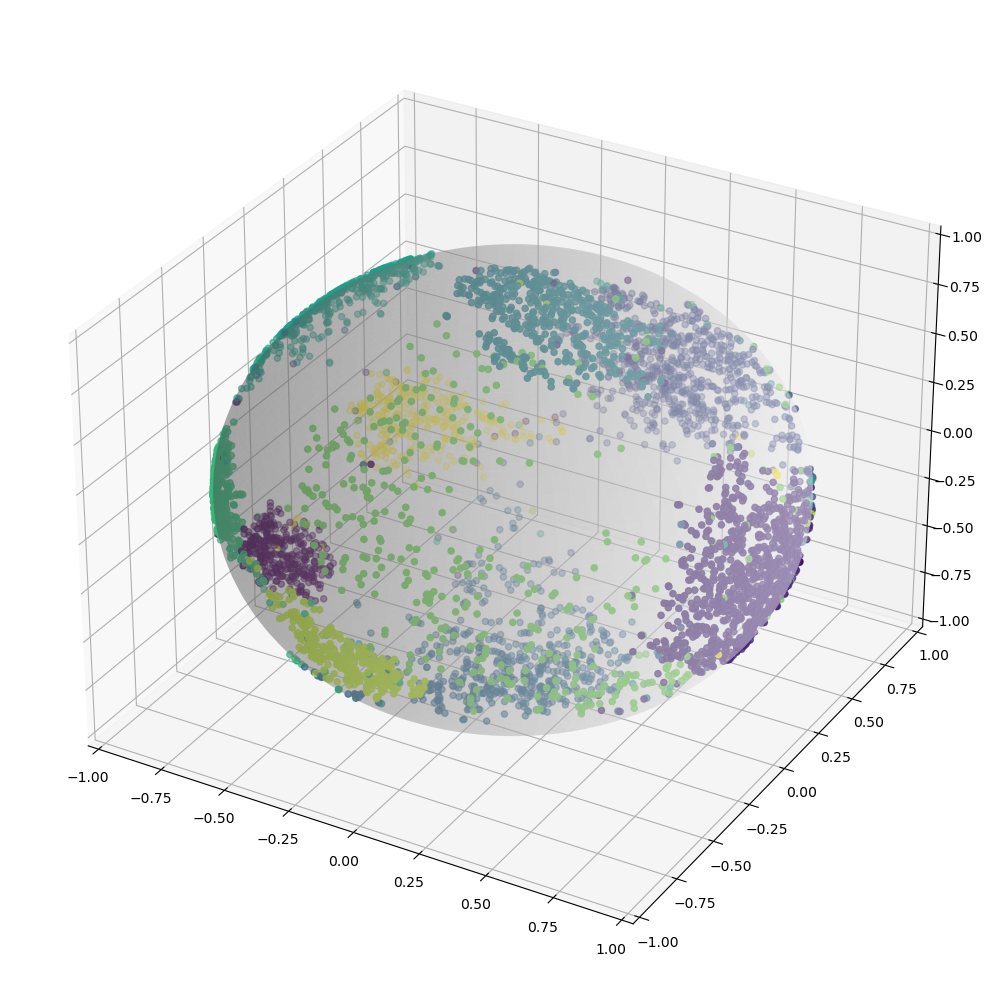}
			\caption{AMC}
			\label{fig:AMC}
		}
	\end{subfigure}
	\begin{subfigure}{.23\textwidth}{
			\includegraphics[width=\textwidth]{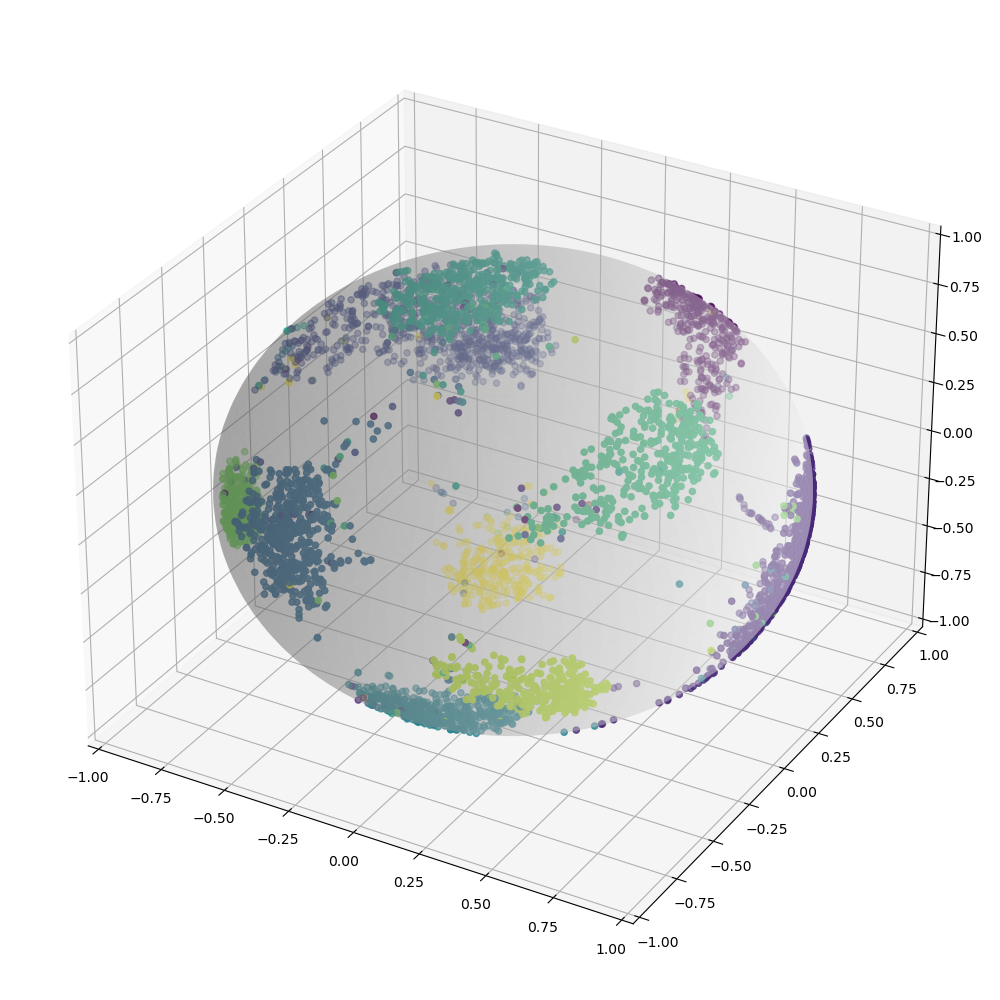}
			\caption{LACE (ours)}
			\label{fig:LACES}
		}
	\end{subfigure}
	\caption{3D t-SNE projection of features extracted from the penultimate layer for each method on the same 5,000 test images from SVHN. The projections are constrained to lie on the unit sphere. Our method improves intra-class compactness and inter-class separation in comparison to all other comparison methods.}
	
	\label{fig:TSNE_visuals} 
\end{figure*}

\begin{table*}[htb!]
\centering
\caption{Cluster validity scores for the Resnet18 backbone on FashionMNIST. The best average result is bolded and the second best result is underlined. As shown in each metric, LACE improves the inter-class separation and intra-class compactness in comparsion to other methods.} 
\begin{tabular}{|c|c|c|c|}
\hline
\vtop{\hbox{\strut Method}\hbox{\strut}} & \vtop{\hbox{\strut Silhouette \cite{rousseeuw1987Silhouette} } \hbox{\strut \hspace{4mm}($1 = $ best)}}  & \vtop{\hbox{\strut Davies-Bouldin \cite{davies1979ClusterValidity} }\hbox{\strut \hspace{4mm}(lowest $ = $ best)}} &\vtop{\hbox{\strut Calinski-Harabasz \cite{calinski1974ClusterValidity} }\hbox{\strut \hspace{4mm}(highest $ = $ best)}} \\ \hline
Softmax & 0.58$\pm$0.01 & 1.07$\pm$0.03 & 3055$\pm$251\\ \hline
Arcface \cite{Deng2019ArcFace} & 0.64$\pm$0.01 & 0.98$\pm$0.00 & 3768$\pm$259\\  \hline
Cosface \cite{Wang2018CosFace} & 0.71$\pm$0.01 & 0.98$\pm$0.05 & 3085$\pm$30\\ \hline 
Sphereface \cite{Liu2017SphereFace} & 0.65$\pm$0.01 & 0.92$\pm$0.01 & 5070$\pm$384\\ \hline
AMC \cite{Choi2020AMCLoss} & 0.71$\pm$0.01 & 0.89$\pm$0.03 & 4703$\pm$122\\ \hline
Center Loss \cite{Wen2016DiscFeatureLearning} & 0.71$\pm$0.01 & 0.98$\pm$0.05 & 3085$\pm$30\\ \hline
LACE (ours) & \underline{0.77$\pm$0.00} & \underline{0.69$\pm$0.00} & \textbf{5737$\pm$195}\\ \hline
LACE (pre-whitened) (ours) & \textbf{0.83$\pm$0.00} & \textbf{0.68$\pm$0.00} & \underline{5640$\pm$174}\\ \hline
\end{tabular}
\label{tab:results_cluster_validity}
\end{table*}

\begin{figure*}[htb]
\centering
	\begin{subfigure}{.45\textwidth}{
			\includegraphics[width=\textwidth]{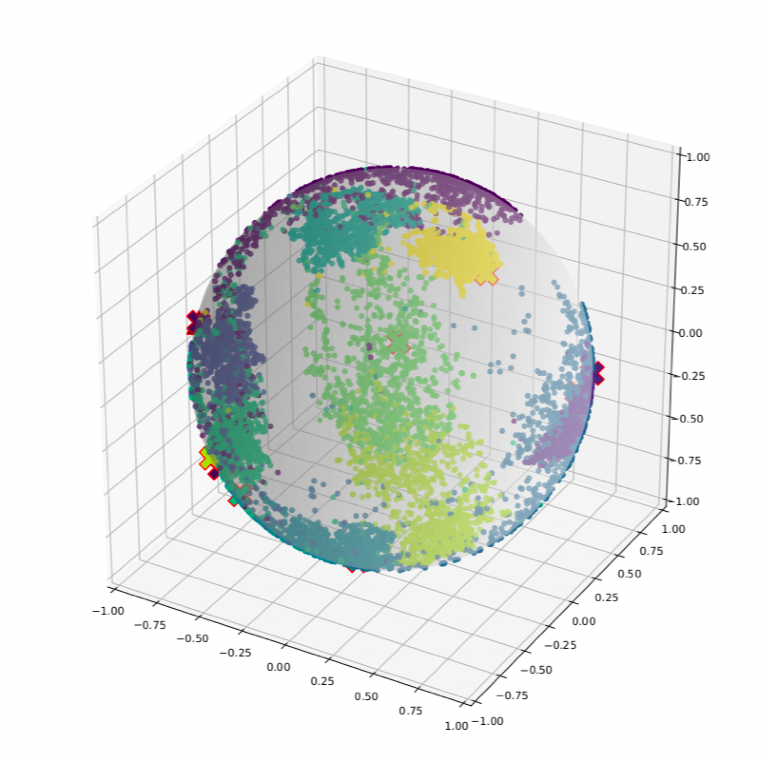}
			\caption{Pre-whitened}
			\label{fig:prewhitened}
		}
	\end{subfigure} 
	\begin{subfigure}{.45\textwidth}{
			\includegraphics[width=\textwidth]{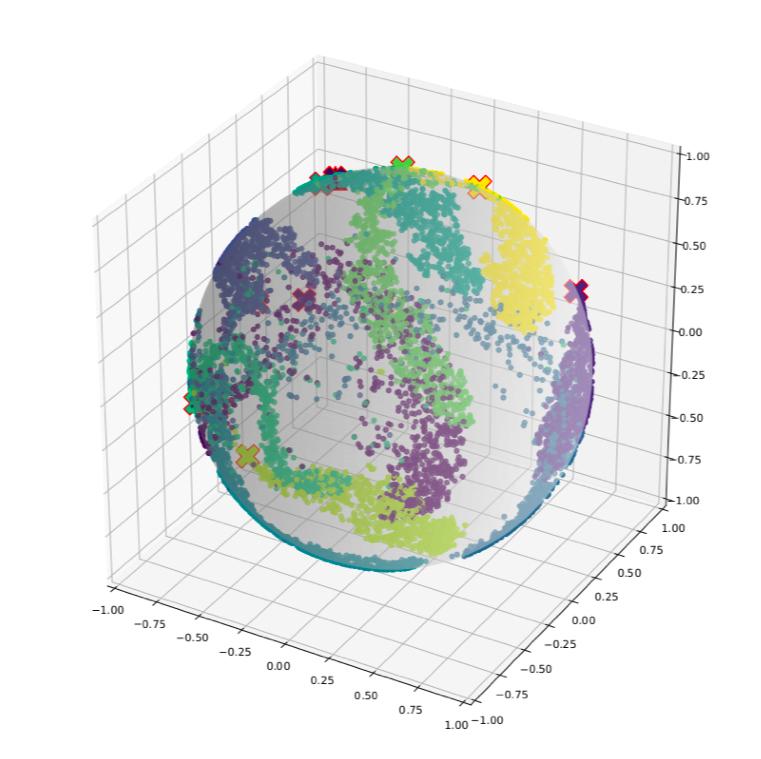}
			\caption{Whitened}
			\label{fig:whitened}
		}
	\end{subfigure}
	\caption{t-SNE visualization of FashionMNIST test data (a) before and (b) after LACE whitening.  Learned target signatures are depicted by `X's outlined in red.  As seen here, data whitening with the learned background statistics reduces feature variance across multiple dimensions. This makes features more separable according to an angular metric, as opposed to Euclidean distance.}
	\label{fig:cluster_validity_whitened_and_prewhitened} 
\end{figure*}

The performances for our proposed approaches and SOA angular softmax/feature regularization methods are shown in Table \ref{tab:CE_outputs}. The results of LACE indicate the effectiveness of the data whitening when comparing against the baseline softmax approach. We can infer that that the background statistics improve discriminability between target and non-target samples. For CIFAR10, FashionMNIST, and SVHN, the LACE approach overall achieved statistically significant performance improvements in comparison to the baseline softmax models. LACE also outperforms the other angular softmax alternatives. Additionally, LACE outperforms feature regularization approaches based on Euclidean (\textit{i.e.}, center loss) and angular (\textit{i.e.}, AMC) information. As a results, we can optimize for only a single objective term that will jointly maximize classification performance and feature regularization.

For CIFAR100 (coarse and sparse), our proposed method did not perform as well in comparison to other SOA approaches. A possible reason for this is that there are a total of $20$ or $100$ classes in this dataset. As a result, it will be difficult to learn background statistics that improve performance for all classes. We can also verify this observation in that the performance of LACE significantly decreases once the number of classes increases from $20$ to $100$ for CIFAR100. For the remaining three datasets, each was comprised of only 10 classes. LACE could be further improved by modeling additional background statistics to account for the various classes. 

The qualitative results for each method on SVHN test data are shown in Figure \ref{fig:TSNE_visuals}. We projected the features from the original $\mathbb{R}^{512}$ to $\mathbb{R}^{3}$ using t-SNE with Cosine as the distance metric. LACE improves the angular class compactness and separation in comparison to the other approaches. For softmax in Figure \ref{fig:Softmax}, there is no direct component in the objective to consider intra-class and inter-class compactness. We observe the same here in the visualization of the features from t-SNE. Center loss, as shown in Figure \ref{fig:Center}, improves intra-class compactness, but there is still considerable overlap with data samples from different classes in the projected space (the classes maybe more separable in higher dimensions). For LACE and other approaches that consider angular information, the features appear more separable and compact with LACE achieving the minimal intra-class and maximum inter-class variation. We further evaluate the discriminitative power of the features learned by LACE through cluster validity metrics. 
\begin{table*}[htb!]
\centering
\caption{Test classification accuracy for each method with different backbone architectures on CIFAR10. The best average result is bolded and the second best result is underlined.} 
\begin{tabular}{|c|c|c|c|c|}
\hline
Method     & ResNet50 \cite{He2015ResNet}        & ResNeXt-50 32x4d \cite{xie2017aggregated}   & WideResNet50-2 \cite{zagoruyko2016wide}   & DenseNet121 \cite{huang2017densely} \\ \hline
Softmax    &89.86$\pm$0.45 & \textbf{91.29$\pm$0.83} & 89.96$\pm$0.17 &  90.82$\pm$0.27 \\ \hline
Arcface \cite{Deng2019ArcFace}    & \underline{89.88$\pm$0.40} & 90.51$\pm$0.29 & 89.66$\pm$0.36 & 90.86$\pm$0.31 \\ \hline
Cosface \cite{Wang2018CosFace}    & \textbf{90.16$\pm$0.23} & \underline{90.86$\pm$0.36} & 89.76$\pm$0.25 & \textbf{91.05$\pm$0.40} \\ \hline 
Sphereface \cite{Liu2017SphereFace}  & 89.58$\pm$0.28 & 90.57$\pm$0.53 & \textbf{90.55$\pm$0.35} & 90.67$\pm$0.55 \\ \hline
AMC \cite{Choi2020AMCLoss} & 89.21$\pm$0.44 & 90.71$\pm$0.50 & 89.37$\pm$0.36 & 90.95$\pm$0.38 \\ \hline
Center Loss \cite{Wen2016DiscFeatureLearning} & 89.77$\pm$0.37 & 89.75$\pm$1.03 & \underline{90.18$\pm$0.47} & \underline{91.00$\pm$0.29} \\ \hline
LACE (ours)     & 87.76$\pm$0.14 & 87.06$\pm$1.46 & 85.51$\pm$0.43 & 84.49$\pm$0.37  \\ \hline
\end{tabular}
\label{tab:backbone}
\end{table*}
\subsection{Clustering Performance}
SOA approaches in the literature aim to improve softmax loss by either reducing intra-class scatter or increasing between-class separation.  In this work, five cluster validity metrics were used to compare the induced distributions of classes.  Namely, Silhouette \cite{rousseeuw1987Silhouette}, Davies-Bouldin (DBI) \cite{davies1979ClusterValidity} and Calinski-Harabasz (also called Variance Ratio Criterion) \cite{calinski1974ClusterValidity} scores were computed to quantify both inter-cluster separation and within-cluster compactness, while homogeneity and completeness \cite{rosenberg2007HomogeneityCompleteness} measured the accuracy of the induced clustering.  Cluster validity scores for each algorithm on FashionMNIST are shown in Table \ref{tab:results_cluster_validity}. As can be seen, LACE features (both whitened and pre-whitened) consistently outperformed the alternatives for the three metrics measuring intra-class compactness and inter-class separation. Homogeneity and completeness essentially measure the label accuracy of samples assigned to clusters.  Performance for these two metrics was very close across all approaches ($0.98$), so the results are negligible. Thus, the networks trained with LACE loss not only obtained the best overall classification accuracy, but also discovered feature representations whose class distributions were the most compact and separable.  

An interesting observation from Table \ref{tab:results_cluster_validity} is that the cluster validity scores on LACE features were better before whitening.  This result might be explained by analysis of Figure \ref{fig:cluster_validity_whitened_and_prewhitened}, which shows t-SNE visualizations of FashionMNIST test points before and after whitening.  The figure shows that whitening with the learned background statistics makes the induced clusters less circular.  This results in a trade-off where within-class compactness is lost, according to the chosen metrics, but between class separation is improved along the angular dimension.  Additionally, in the whitened space, cluster variance is increased along an angle but decreased in two other dimensions.  Thus, the LACE whitening might sacrifice the traditional ideal of projecting all data of the same class onto a single point to improve discriminability in the angular sense.  This result matches intuition of classification with ACE.

\subsection{Ablation Study}
\label{sec:Ablation}
\paragraph{Backbone Architectures}
To further evaluate our proposed method, we investigate applying different backbone architectures to learn and extract features of varying dimensionality. We used four additional backbone architectures in addition to ResNet18: ResNet50 \cite{He2015ResNet}, ResNeXt-50 32x4d \cite{xie2017aggregated}, WideResNet50-2 \cite{zagoruyko2016wide}, and DenseNet121 \cite{huang2017densely}. The feature dimensionality of ResNet50, ResNeXt-50 32x4d, and WideResNet50-2 was $\mathbb{R}^{2048}$ while DenseNet121 was $\mathbb{R}^{1024}$. We present results of these experiments for CIFAR10 in Table \ref{tab:backbone}. From the results, we observe that performance is stable (\textit{i.e.}, small standard deviations) across all methods. The results demonstrate the robustness of our method across different backbone architectures. LACE can thus be integrated seamlessly into various deep learning models to learn discriminative features to improve intra-class compactness and inter-class separation.

In comparison to Table \ref{tab:CE_outputs}, the results of the other methods improved while LACE performed slightly below the previous results with ResNet18. For each of the new backbone architectures, the feature dimensionality increased in comparison to ResNet18 ($\mathbb{R}^{512}$). As stated previously, it will become more difficult to learn global background statistics that optimize classification performance for all classes as dimensionality increases. Another observation is that LACE performs best with a smaller (\textit{i.e.}, fewer parameters) model. As a result, models in future work could potentially use LACE as a method to maintain and/or improve performance while reducing the number of parameters needed by deeper artificial neural networks.

\paragraph{Impact of LACE components}

\begin{table}[b!]
\centering
\caption{We evaluated the components of LACE that made the most impact on performance by using CIFAR10: background mean ($\mu_b$) and background covariance ($\Sigma_b$). The features and target signature matrix were $L2$ normalized for each combination. A \checkmark indicates that component was active during the experimental runs.}
\begin{tabular}{|c|c|c|}
\hline
$\mu_b$                   & $\Sigma_b$              & CIFAR10 \\ \hline
&    &     85.49$\pm$3.21     \\ \hline
& \checkmark &   87.93$\pm$0.14  \\ \hline
\checkmark &  &   87.26$\pm$1.54 \\ \hline
\checkmark & \checkmark & \textbf{90.26$\pm$0.19}     \\ \hline
\end{tabular}
\label{tab:ablation}
\end{table}
In order to provide more insight into our proposed method, we performed an ablation study of the different components of our approach. LACE is comprised of two major components: the background mean ($\mu_b$) and background covariance ($\Sigma_b$). We evaluated each component of LACE for the CIFAR10 dataset with ResNet18 as the backbone architecture and the test accuracy is shown in Table \ref{tab:ablation}. We observe that with only $L2$ normalization, the performance is not statistically significant in comparison to the baseline softmax results in Table \ref{tab:CE_outputs}. The typical cross entropy loss does not directly account for inter-class relationships in the data. By including the background statistics in our loss, we can increase the separation between data samples in different classes and map samples in the same class closer to one another. 

Including the background covariance improved the performance of the baseline model in conjunction with the $L2$ normalization. The model is able to learn meaningful rotations of the data that can account for both inter-class and intra-class relationships to maximize classification performance. The background mean also improved performance of baseline results as well. The background mean serves as a way to shift the data as we saw in the toy example in Figure \ref{fig:neural_ace_loss_network}. The background mean provides additional flexibility to the model. Using the background mean and covariance separately lead to similar performance with the background covariance achieving a slightly higher average test accuracy. Since artificial neural networks are inherently learning these ``angular" feature distributions, the background covariance matrix will be the most powerful component of the whitened data transformation in LACE. When both background statistics are used, we achieve optimal performance for CIFAR10.

\subsection{Computational Cost of LACE}
LACE has shown to improve performance, but we must also consider the additional cost of the proposed method. In comparison to the traditional softmax output layer, LACE introduces two additional components: background mean and background covariance. By adding the background statistics, we introduce $d(d+1)$ parameters to be learned ($d$ and $d^2$ for the background mean and covariance respectively). As the feature dimensionality increases, the computational costs of the whitening operation will increase. However, as shown in other whitening works \cite{Su2021Whitening,huang2019iterative,zhang2021stochastic}, various optimization approaches can be used to improve the cost of computing the background statistics of the proposed method.

\section{Conclusion}
We presented a new loss function, LACE, that can be used to improve image classification performance. Our proposed method uses the angular features inherently learned by artificial neural networks, tuning them to become more discriminative data representations. Qualitative and quantitative analysis show that LACE outperforms and/or performs comparatively with other SOA angular softmax and feature regularization approaches. Additionally, LACE does not need an angular margin and/or scale set \textit{a priori} since all parameters are learned during training.

Our proposed approach outperforms or achieves comparable performance to alternative SOA angular margin approaches in several benchmark image classification tasks.  A notable observation is that LACE typically achieves better performance when implemented with smaller models and fewer classes.  To improve performance with larger models, several approaches could be investigated.  Instead of assuming a global background distribution, the authors believe performance can be further improved through one-vs-all or one-vs-one classification where each class has it's own corresponding background mean and covariance. Additionally, alternative update approaches could be implemented to update the LACE parameters (background means, background covariance, target signatures). Future work will explore parameter estimation from the data, directly, as well as employing physical constraints on the possible solutions to adhere more closely to the data.  This could be beneficial for learning both discriminative and physically meaningful representations.  Finally, as with batch normalization, LACE may be applied at different portions of the deep learning model.  Future investigation will explore how whitening with background statistics within the model affect classification performance and training stability. 
\label{sec:conclusion}

\section*{Acknowledgments}
This material is based upon work supported by the National Science Foundation Graduate Research Fellowship under Grant No. DGE-1842473. The views and opinions of authors expressed herein do not necessarily state or reflect those of the United States Government or any agency thereof.

\bibliographystyle{ieee_fullname}
\bibliography{references}

\section*{Appendix}
\appendix
\counterwithin*{equation}{section}
We provide the derivations for the derivatives of the parameters for LACE as discussed in Section \ref{sec:lace_loss}. The derivatives for the target signatures is shown in Section \ref{sect:target} and the background statistics (\textit{i.e.}, mean and covariance) are shown in Sections \ref{sec:background_mean} and \ref{sec:background_cov}.
\onecolumn
\section{Target Signatures} \label{sect:target}

\begin{equation}
      L_{LACE} = \frac{1}{B}\sum_{n=1}^{B}-\log \left ( \frac{\exp{\doublehat{\bm{s}}_{c}^T\doublehat{\bm{x}_n}}}{{\sum_{j=1}^{C}}\exp{\doublehat{\bm{s}}}_{j}^T\doublehat{\bm{x}_n}} \right) 
\end{equation}

\begin{equation}
      L_{LACE} = \frac{-1}{B}\sum_{n=1}^{B} \left ( \doublehat{\bm{s}}_{c}^T\doublehat{\bm{x}_n}-\log \left({\sum_{j=1}^{C}}\exp{\doublehat{\bm{s}}}_{j}^T\doublehat{\bm{x}_n} \right) \right) 
\end{equation}

\begin{equation}
      L_{LACE} = \frac{-1}{B}\sum_{n=1}^{B} \left ( \doublehat{\bm{x}_n}^T\doublehat{\bm{s}}_{c}-\log \left({\sum_{j=1}^{C}}\exp{\doublehat{\bm{x}_n}^T\doublehat{\bm{s}}}_{j}\right) \right) 
\end{equation}

\begin{equation}
      L_{LACE} = \frac{-1}{B}\sum_{n=1}^{B} \left ( \doublehat{\bm{x}_n}^T\left(\frac{\bm{D}^{-\frac{1}{2}}\bm{U}^{T}\bm{s}_c}{\lVert \bm{D}^{-\frac{1}{2}}\bm{U}^{T}\bm{s}_c \rVert}\right)-\log \left({\sum_{j=1}^{C}}\exp{\doublehat{\bm{x}_n}^T\left(\frac{\bm{D}^{-\frac{1}{2}}\bm{U}^{T}\bm{s}_j}{\lVert \bm{D}^{-\frac{1}{2}}\bm{U}^{T}\bm{s}_j \rVert}\right)}\right) \right) 
\end{equation}

\begin{equation}
\bm{z}_{c} = \bm{D}^{-\frac{1}{2}}\bm{U}^{T}\bm{s}_c
\end{equation}

\begin{equation}
      L_{LACE} = \frac{-1}{B}\sum_{n=1}^{B} \left ( \doublehat{\bm{x}_n}^T\left(\frac{\bm{z}_{c}}{\lVert\bm{z}_{c} \rVert}\right)-\log \left({\sum_{j=1}^{C}}\exp{\doublehat{\bm{x}_n}^T\left(\frac{\bm{z}_{j}}{\lVert\bm{z}_{j} \rVert}\right)}\right) \right) 
\end{equation}

\begin{equation}
\bm{z'}_{c} = \cfrac{\partial}{\partial \bm{s}_{c}} \left(\frac{\bm{z}_{c}}{\lVert\bm{z}_{c} \rVert}\right)= \cfrac{\partial}{\partial \bm{s}_{c}}\left(\frac{\bm{D}^{-\frac{1}{2}}\bm{U}^{T}\bm{s}_c}{\lVert \bm{D}^{-\frac{1}{2}}\bm{U}^{T}\bm{s}_c \rVert}\right)=\left(\cfrac{\lVert \bm{z}_{c} \rVert(\bm{D}^{-\frac{1}{2}}\bm{U}^{T}) - \cfrac{\bm{z}_{c}\left(\bm{U}\bm{D}^{-1}\bm{U}^{T}\bm{s}_{c}\right)^T}{\lVert \bm{z}_{c} \rVert}}{\lVert \bm{z}_{c} \rVert^2}\right)
\end{equation}

\begin{equation}
      \frac{\partial L_{LACE}}{\partial \bm{s}_{c}} = \frac{-1}{B}\sum_{n=1}^{B} \frac{\partial}{\partial \bm{s}_{c}}\left ( \doublehat{\bm{x}_n}^T\left(\frac{\bm{z}_{c}}{\lVert\bm{z}_{c} \rVert}\right)-\log \left({\sum_{j=1}^{C}}\exp{\doublehat{\bm{x}_n}^T\left(\frac{\bm{z}_{j}}{\lVert\bm{z}_{j} \rVert}\right)}\right) \right)
\end{equation}

\begin{equation}
      \frac{\partial L_{LACE}}{\partial \bm{s}_{c}} = \frac{-1}{B}\sum_{n=1}^{B}\left ( \doublehat{\bm{x}_n}^T\bm{z'}_{c}-\cfrac{\exp\doublehat{\bm{x}_n}^T\left(\frac{\bm{z}_{c}}{\lVert\bm{z}_{c} \rVert}\right)\doublehat{\bm{x}_n}^T\bm{z'}_c} {{\sum_{j=1}^{C}}\exp{\doublehat{\bm{x}_n}^T\left(\frac{\bm{z}_{j}}{\lVert\bm{z}_{j} \rVert}\right)}} \right)
\end{equation}

\section{Background Mean}\label{sec:background_mean}
\begin{equation}
      L_{LACE} = \frac{1}{B}\sum_{n=1}^{B}-\log \left ( \frac{\exp{\doublehat{\bm{s}}_{c}^T\doublehat{\bm{x}_n}}}{{\sum_{j=1}^{C}}\exp{\doublehat{\bm{s}}}_{j}^T\doublehat{\bm{x}_n}} \right) 
\end{equation}

\begin{equation}
      L_{LACE} = \frac{-1}{B}\sum_{n=1}^{B} \left ( \doublehat{\bm{s}}_{c}^T\doublehat{\bm{x}_n}-\log \left({\sum_{j=1}^{C}}\exp{\doublehat{\bm{s}}}_{j}^T\doublehat{\bm{x}_n} \right) \right) 
\end{equation}

\begin{equation}
      L_{LACE} = \frac{-1}{B}\sum_{n=1}^{B} \left ( \doublehat{\bm{s}}_{c}^T\left(\frac{\bm{D}^{-\frac{1}{2}}\bm{U}^{T}(\bm{x}_n-\bm{\mu}_{b})}{\lVert\bm{D}^{-\frac{1}{2}}\bm{U}^{T}(\bm{x}_n-\bm{\mu}_{b})\rVert}\right)-\log \left({\sum_{j=1}^{C}}\exp{\doublehat{\bm{s}}}_{j}^T\left(\frac{\bm{D}^{-\frac{1}{2}}\bm{U}^{T}(\bm{x}_n-\bm{\mu}_{b})}{\lVert\bm{D}^{-\frac{1}{2}}\bm{U}^{T}(\bm{x}_n-\bm{\mu}_{b})\rVert}\right)\right) \right) 
\end{equation}

\begin{equation}
\bm{m}_{b} = \bm{D}^{-\frac{1}{2}}\bm{U}^{T}(\bm{x}_n-\bm{\mu}_{b})
\end{equation}

\begin{equation}
      L_{LACE} = \frac{-1}{B}\sum_{n=1}^{B} \left ( \doublehat{\bm{s}}_{c}^T\left(\frac{\bm{m}_b}{\lVert\bm{m}_b\rVert}\right)-\log \left({\sum_{j=1}^{C}}\exp{\doublehat{\bm{s}}}_{j}^T\left(\frac{\bm{m}_b}{\lVert\bm{m}_b\rVert}\right)\right) \right) 
\end{equation}

\begin{equation}
      \frac{\partial L_{LACE}}{\partial \bm{\mu}_{b}} = \frac{-1}{B}\sum_{n=1}^{B} \frac{\partial}{\partial \bm{\mu}_{b}}\left ( \doublehat{\bm{s}}_{c}^T\left(\frac{\bm{m}_b}{\lVert\bm{m}_b\rVert}\right)-\log \left({\sum_{j=1}^{C}}\exp{\doublehat{\bm{s}}}_{j}^T\left(\frac{\bm{m}_b}{\lVert\bm{m}_b\rVert}\right)\right) \right) 
\end{equation}

\begin{equation}
\bm{m'}_{b} = \cfrac{\partial}{\partial \bm{\mu}_{b}}\left(\cfrac{\bm{m}_b}{\lVert\bm{m}_b\rVert}\right)=\cfrac{\partial}{\partial \bm{\mu}_{b}}\left(\cfrac{\bm{D}^{-\frac{1}{2}}\bm{U}^{T}(\bm{x}_n-\bm{\mu}_{b})}{\lVert\bm{D}^{-\frac{1}{2}}\bm{U}^{T}(\bm{x}_n-\bm{\mu}_{b})\rVert}\right)=\cfrac{-\lVert\bm{m}_b\rVert\bm{D}^{-\frac{1}{2}}\bm{U}^{T}-\cfrac{\bm{m}_{b}\left(\bm{x}_n^{T}\Sigma_b^{-1}+\Sigma_b^{-1}\bm{\mu}_b\right)}{\lVert \bm{m}_{b} \rVert}}{\lVert\bm{m}_b\rVert^2}
\end{equation}

\begin{equation}
      \frac{\partial L_{LACE}}{\partial \bm{\mu}_{b}} = \frac{-1}{B}\sum_{n=1}^{B}\left ( \doublehat{\bm{s}}_{c}^T\bm{m'}_{b}-\cfrac{{\sum_{j=1}^{C}}\exp{\doublehat{\bm{s}}}_{j}^T\left(\frac{\bm{m}_b}{\lVert\bm{m}_b\rVert}\right)\doublehat{\bm{s}}_j^T\bm{m'}_b} {{\sum_{j=1}^{C}}\exp{\doublehat{\bm{s}}}_{j}^T\left(\frac{\bm{m}_b}{\lVert\bm{m}_b\rVert}\right)} \right) 
\end{equation}

\begin{equation}
      \frac{\partial L_{LACE}}{\partial \bm{\mu}_{b}} = \frac{-1}{B}\sum_{n=1}^{B}\left ( \doublehat{\bm{s}}_{c}^T\bm{m'}_{b}-\sum_{j=1}^{C}\doublehat{\bm{s}}_j^T\bm{m'}_{b}\right) 
\end{equation}

\begin{equation}
      \frac{\partial L_{LACE}}{\partial \bm{\mu}_{b}} = \frac{1}{B}\sum_{n=1}^{B}\underset{j \neq c}{\sum_{j=1}^{C}}\doublehat{\bm{s}}_j^T\bm{m'}_{b}
\end{equation}

\begin{equation}
      \frac{\partial L_{LACE}}{\partial \bm{\mu}_{b}} = \frac{1}{B}\sum_{n=1}^{B}\underset{j \neq c}{\sum_{j=1}^{C}}\doublehat{\bm{s}}_j^T\bm{D}^{-\frac{1}{2}}\bm{U}^{T}(\bm{x}_n-\bm{\mu}_{b})
\end{equation}

\section{Inverse Background Covariance}\label{sec:background_cov}
\textit{Note}: In the code, we do not compute the inverse. Instead, by enforcing this metric, we expect the algorithm to learn a matrix that is the inverse background covariance. We take the derivative here with respect to $\Sigma_{b}^{-1}$.

\begin{equation}
      L_{LACE} = \frac{1}{B}\sum_{n=1}^{B}-\log \left ( \frac{\exp{\doublehat{\bm{s}}_{c}^T\doublehat{\bm{x}_n}}}{{\sum_{j=1}^{C}}\exp{\doublehat{\bm{s}}}_{j}^T\doublehat{\bm{x}_n}} \right) 
\end{equation}

\begin{equation}
      L_{LACE} = \frac{-1}{B}\sum_{n=1}^{B} \left ( \doublehat{\bm{s}}_{c}^T\doublehat{\bm{x}_n}-\log \left({\sum_{j=1}^{C}}\exp{\doublehat{\bm{s}}}_{j}^T\doublehat{\bm{x}_n} \right) \right) 
\end{equation}

\begin{equation}
\doublehat{\bm{s}}_{c}^T\doublehat{\bm{x}_n}=\cfrac{\bm{s}_c^{T}\Sigma_{b}^{-1}(\bm{x}_n-\bm{\mu}_{b})}{\lVert\bm{s}_c^{T}\Sigma_{b}^{-1}(\bm{x}_n-\bm{\mu}_{b})\rVert}
\end{equation}

\begin{equation}
      L_{LACE} = \frac{-1}{B}\sum_{n=1}^{B} \left (\cfrac{\bm{s}_c^{T}\Sigma_{b}^{-1}(\bm{x}_n-\bm{\mu}_{b})}{\lVert\bm{s}_c^{T}\Sigma_{b}^{-1}(\bm{x}_n-\bm{\mu}_{b})\rVert}-\log \left({\sum_{j=1}^{C}}\exp{\cfrac{\bm{s}_j^{T}\Sigma_{b}^{-1}(\bm{x}_n-\bm{\mu}_{b})}{\lVert\bm{s}_j^{T}\Sigma_{b}^{-1}(\bm{x}_n-\bm{\mu}_{b})\rVert}} \right) \right) 
\end{equation}

\begin{equation}
\mathcal{V}'_c=\frac{\partial}{\partial \bm{\Sigma}_{b}^{-1}}\lVert\bm{s}_c^{T}\Sigma_{b}^{-1}(\bm{x}_n-\bm{\mu}_{b})\rVert=\cfrac{\bm{s}_c\bm{s}_c^T\left(\bm{x}_n-\bm{\mu}_{b})^{T}\Sigma_{b}^{-1}(\bm{x}_n-\bm{\mu}_{b}\right)+\bm{s}_c^T\Sigma_{b}^{-1}\bm{s}_c(\bm{x}_n-\bm{\mu}_{b})(\bm{x}_n-\bm{\mu}_{b})^T}{2\lVert\bm{s}_c^{T}\Sigma_{b}^{-1}(\bm{x}_n-\bm{\mu}_{b})\rVert}
\end{equation}

\begin{equation}
V'_{c}=\frac{\partial}{\partial \bm{\Sigma}_{b}^{-1}}\left (\cfrac{\bm{s}_c^{T}\Sigma_{b}^{-1}(\bm{x}_n-\bm{\mu}_{b})}{\lVert\bm{s}_c^{T}\Sigma_{b}^{-1}(\bm{x}_n-\bm{\mu}_{b})\rVert}\right)=
\left (\cfrac{\lVert\bm{s}_c^{T}\Sigma_{b}^{-1}(\bm{x}_n-\bm{\mu}_{b})\rVert(\bm{x}_n-\bm{\mu}_{b})\bm{s}_c^{T}-\bm{s}_c^{T}\Sigma_{b}^{-1}(\bm{x}_n-\bm{\mu}_{b})\mathcal{V}'_c}{\lVert\bm{s}_c^{T}\Sigma_{b}^{-1}(\bm{x}_n-\bm{\mu}_{b})\rVert^2}\right)
\end{equation}

\begin{equation}
      \frac{\partial L_{LACE}}{\partial \bm{\Sigma}_{b}^{-1}} = \frac{-1}{B}\sum_{n=1}^{B} \frac{\partial}{\partial \bm{\Sigma}_{b}^{-1}}\left(\cfrac{\bm{s}_c^{T}\Sigma_{b}^{-1}(\bm{x}_n-\bm{\mu}_{b})}{\lVert\bm{s}_c^{T}\Sigma_{b}^{-1}(\bm{x}_n-\bm{\mu}_{b})\rVert}-\log \left({\sum_{j=1}^{C}}\exp{\cfrac{\bm{s}_j^{T}\Sigma_{b}^{-1}(\bm{x}_n-\bm{\mu}_{b})}{\lVert\bm{s}_j^{T}\Sigma_{b}^{-1}(\bm{x}_n-\bm{\mu}_{b})\rVert}} \right) \right) 
\end{equation}

\begin{equation}
      \frac{\partial L_{LACE}}{\partial \bm{\Sigma}_{b}^{-1}} = \frac{-1}{B}\sum_{n=1}^{B} \left(V'_{c}-\cfrac{{\sum_{j=1}^{C}}\exp{\cfrac{\bm{s}_j^{T}\Sigma_{b}^{-1}(\bm{x}_n-\bm{\mu}_{b})}{\lVert\bm{s}_j^{T}\Sigma_{b}^{-1}(\bm{x}_n-\bm{\mu}_{b})\rVert}}}{{\sum_{j=1}^{C}}\exp{\cfrac{\bm{s}_j^{T}\Sigma_{b}^{-1}(\bm{x}_n-\bm{\mu}_{b})}{\lVert\bm{s}_j^{T}\Sigma_{b}^{-1}(\bm{x}_n-\bm{\mu}_{b})\rVert}}}V'_{j} \right) 
\end{equation}

\begin{equation}
      \frac{\partial L_{LACE}}{\partial \bm{\Sigma}_{b}^{-1}} = \frac{-1}{B}\sum_{n=1}^{B} \left(V'_{c}-\sum_{j=1}^{C}V'_{j} \right) 
\end{equation}

\begin{equation}
      \frac{\partial L_{LACE}}{\partial \bm{\Sigma}_{b}^{-1}} = \frac{1}{B}\sum_{n=1}^{B}\underset{j \neq c}{\sum_{j=1}^{C}}V'_{j}
\end{equation}

\small
\begin{equation}
      \frac{\partial L_{LACE}}{\partial \bm{\Sigma}_{b}^{-1}} = \frac{1}{B}\sum_{n=1}^{B}\underset{j \neq c}{\sum_{j=1}^{C}}\left (\cfrac{\lVert\bm{s}_j^{T}\Sigma_{b}^{-1}(\bm{x}_n-\bm{\mu}_{b})\rVert(\bm{x}_n-\bm{\mu}_{b})\bm{s}_j^{T}-\bm{s}_j^{T}\Sigma_{b}^{-1}(\bm{x}_n-\bm{\mu}_{b})\mathcal{V}'_j}{\lVert\bm{s}_j^{T}\Sigma_{b}^{-1}(\bm{x}_n-\bm{\mu}_{b})\rVert^2}\right)
\end{equation}
\end{document}